\documentclass[10pt,twocolumn,letterpaper]{article}

\usepackage{cvpr}
\usepackage{times}
\usepackage{epsfig}
\usepackage{graphicx}
\usepackage[labelsep=period]{caption}
\usepackage{subcaption}
\usepackage{tabularx}
\usepackage{multirow}
\usepackage{amsmath}
\usepackage{amssymb}
\usepackage[nocompress]{cite}

\usepackage[pagebackref=true,breaklinks=true,letterpaper=true,colorlinks,bookmarks=false]{hyperref}

\cvprfinalcopy 

\def\cvprPaperID{1262} 

\ifcvprfinal\fi
\begin{document}

\title{Joint Detection and Identification Feature Learning for Person Search}

\author{Tong Xiao$^{1*}$\ \ \ \ Shuang Li$^{1}$\thanks{Tong Xiao and Shuang Li are co-first authors with equal contributions.}\ \ \ \ Bochao Wang$^{2}$\ \ \ \ Liang Lin$^{2,3}$\ \ \ \ Xiaogang Wang$^{1}$\\
\small $^{1}$The Chinese University of Hong Kong~~\small $^{2}$Sun Yat-Sen University~~\small $^{3}$SenseTime Group Limited\\
{\tt\small \{xiaotong,sli,xgwang\}@ee.cuhk.edu.hk,\ wangboch@mail2.sysu.edu.cn,\ linliang@ieee.org}
}

\maketitle
\thispagestyle{empty}

\begin{abstract}
Existing person re-identification benchmarks and methods mainly focus on matching cropped pedestrian images between queries and candidates. However, it is different from real-world scenarios where the annotations of pedestrian bounding boxes are unavailable and the target person needs to be searched from a gallery of whole scene images. To close the gap, we propose a new deep learning framework for person search. Instead of breaking it down into two separate tasks---pedestrian detection and person re-identification, we jointly handle both aspects in a single convolutional neural network. An Online Instance Matching (OIM) loss function is proposed to train the network effectively, which is scalable to datasets with numerous identities. To validate our approach, we collect and annotate a large-scale benchmark dataset for person search. It contains $18,184$ images, $8,432$ identities, and $96,143$ pedestrian bounding boxes. Experiments show that our framework outperforms other separate approaches, and the proposed OIM loss function converges much faster and better than the conventional Softmax loss.
\end{abstract}

\section{Introduction} 
\label{sec:introduction}
Person re-identification (re-id)~\cite{zajdel2005keeping,gheissari2006person} aims at matching a target person with a gallery of pedestrian images. It has many video surveillance applications, such as finding criminals~\cite{wang2013intelligent}, cross-camera person tracking~\cite{yu2013harry}, and person activity analysis~\cite{loy2009multi}. The problem is challenging because of complex variations of human poses, camera viewpoints, lighting, occlusion, resolution, background clutter, \etc, and thus draws much research attention in recent years~\cite{zheng2015scalable,liao2015person,paisitkriangkrai2015learning,xiao2016learning,li2014deepreid,chu2016structured}.

\begin{figure}[t]
\begin{center}
\begin{subfigure}[b]{\linewidth}
   \includegraphics[width=\linewidth]{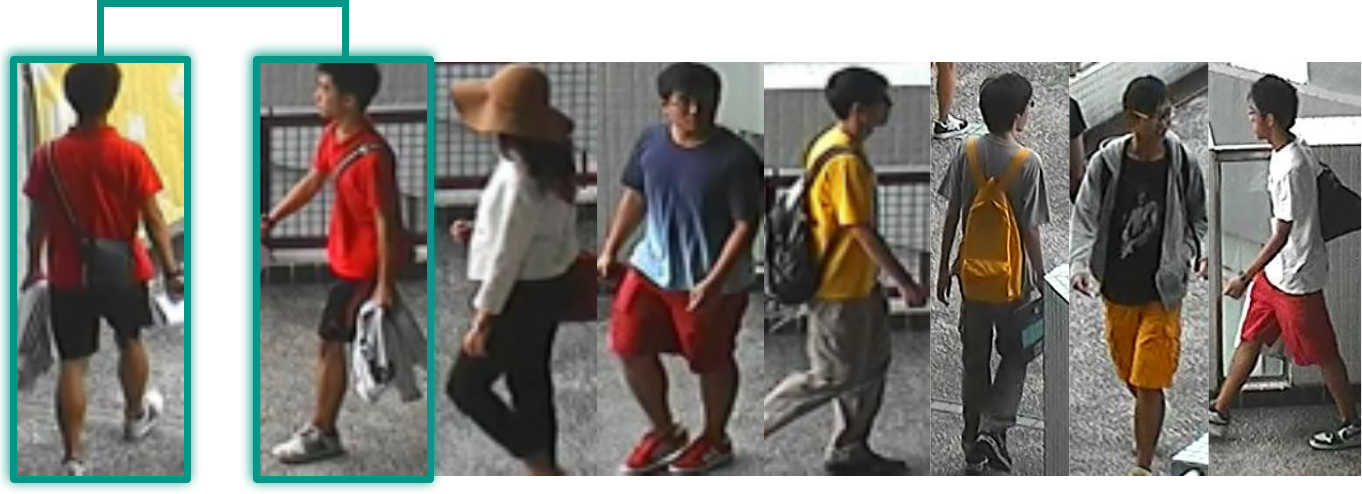}
   \caption{Person re-id: matching with manually cropped pedestrians}
   \label{fig:intro-person-reid}
\end{subfigure}
\\[1em]
\begin{subfigure}[b]{\linewidth}
   \includegraphics[width=\linewidth]{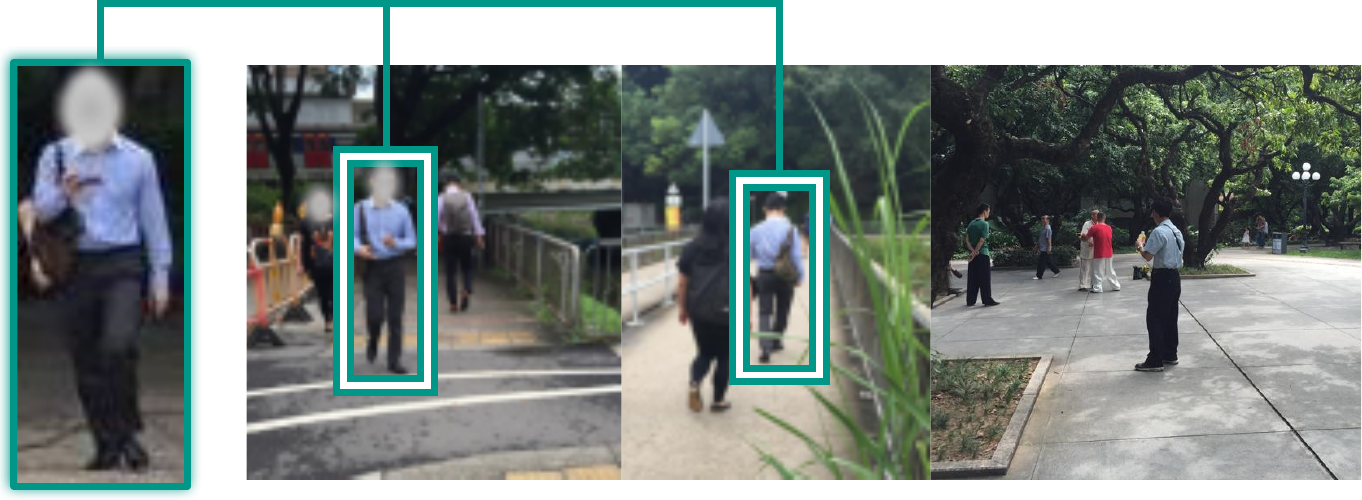}
   \caption{Person search: finding from whole scene images}
   \label{fig:intro-person-search}
\end{subfigure}
\end{center}
\vspace{-1em}\caption{Comparison between person re-identification and person search. The person search problem setting is closer to real-world applications and more challenging, as detecting pedestrians would inevitably produce false alarms, misdetections, and misalignments.}
\label{fig:intro}
\end{figure}

Although numerous person re-id datasets and methods have been proposed, there is still a big gap between the problem setting itself and real-world applications. In most benchmarks~\cite{li2014deepreid,gray2007evaluating,hirzer2011person,li2013locally,zheng2009associating}, the gallery only contains manually cropped pedestrian images (Figure~\ref{fig:intro-person-reid}), while in real applications, the goal is to find a target person in a gallery of whole scene images, as shown in Figure~\ref{fig:intro-person-search}. Following the protocols of these benchmarks, most of the existing person re-id methods assume perfect pedestrian detections. However, these manually cropped bounding boxes are unavailable in practical applications. Off-the-shelf pedestrian detectors would inevitably produce false alarms, misdetections, and misalignments, which could harm the final searching performance significantly.

In 2014, Xu~\etal~\cite{xu2014person} made the first step towards closing this gap. They introduced the person search problem to the community, and proposed a sliding window searching strategy based on a combination of pedestrian detection and person matching scores. However, the performance is limited by the handcrafted features, and the sliding window framework is not scalable.

In this paper, we propose a new deep learning framework for person search. Different from conventional approaches that break down the problem into two separate tasks---pedestrian detection and person re-identification, we jointly handle both aspects in a single Convolutional Neural Network (CNN). Our CNN consists of two parts, given a whole input gallery image, a pedestrian proposal net is used to produce bounding boxes of candidate people, which are fed into an identification net to extract features for comparing with the target person. The pedestrian proposal net and the identification net adapt with each other during the joint optimization. For example, the proposal net can focus more on the recall rather than the precision, as false alarms could be eliminated through the latter features matching process. Meanwhile, misalignments of proposals are also acceptable, as they can be further adjusted by the identification net. To improve the scalability of the whole system, inspired by recent advances in object detection~\cite{ren2015faster}, we encourage both parts to share underlying convolutional feature maps, which significantly accelerates the inference procedure.

Traditional re-id feature learning mainly employs pairwise or triplet distance loss functions~\cite{li2014deepreid,ahmed2015improved,cheng2016person,ding2015deep}. However, they are not efficient as only several data samples are compared at each time, and there are $O(N^2)$ potential input combinations, where $N$ is the number of images. Different sampling strategies could significantly impact the convergence rate and quality, but finding efficient sampling strategies becomes much more difficult as $N$ increases. Another approach is learning to classify identities with the Softmax loss function~\cite{xiao2016learning}, which effectively compares all the samples at the same time. But as the number of classes increases, training the big Softmax classifier matrix becomes much slower or even cannot converge. In this paper, we propose a novel \textit{Online Instance Matching (OIM)} loss function to cope with the problems. We maintain a lookup table of features from all the labeled identities, and compare distances between mini-batch samples and all the registered entries. On the other hand, many unlabeled identities could appear in scene images, which can be served as negatives for labeled identities. We thus exploit a circular queue to store their features also for comparison. This is another advantage brought by the person search problem setting. The proposed parameter-free OIM loss converges much faster and better than the Softmax loss in our experiments.

The contribution of our work is three-fold. First, we propose a new deep learning framework to search a target person from a gallery of whole scene images. Instead of simply combining the pedestrian detectors and person re-id methods, we jointly optimize both objectives in a single CNN and they better adapt with each other. Second, we propose an Online Instance Matching loss function to learn identification features more effectively, which enables our framework to be scalable to large datasets with numerous identities. Together with the fast inference speed, our framework is much closer to the real-world application requirements. At last, we collect and annotate a large-scale benchmark dataset for person search, covering hundreds of scenes from street and movie snapshots. The dataset contains $18,184$ images, $8,432$ identities, and $96,143$ pedestrian bounding boxes. We validate the effectiveness of our approach comparing against other baselines on this dataset. The dataset and code are made public to facilitate further research\footnote{\url{https://github.com/ShuangLI59/person_search}}.

\section{Related Work} 
\label{sec:related_work}
\textbf{Person re-identification.} Early person re-identification methods addressed the problem by manually designing discriminative features~\cite{wang2007shape,hamdoun2008person,zhao2013unsupervised}, learning feature transforms across camera views~\cite{prosser2010person,porikli2003inter,shen2015person}, and learning distance metrics~\cite{zheng2011person,gray2008viewpoint,prosser2010person,paisitkriangkrai2015learning,liao2015efficient}. Recent years, many researchers have proposed various deep learning based methods that jointly handle all these aspects. Li~\etal~\cite{li2014deepreid} and Ahmed~\etal~\cite{ahmed2015improved} designed specific CNN models for person re-id. Both the networks utilize as input a pair of cropped pedestrian images and employ a binary verification loss function to train the parameters. Ding~\etal~\cite{ding2015deep} and Cheng~\etal~\cite{cheng2016person} exploited triplet samples for training CNNs to minimize the feature distance between the same person and maximize the distance between different people. Apart from using pairwise or triplet loss functions, Xiao~\etal~\cite{xiao2016learning} proposed to learn features by classifying identities. Multiple datasets are combined together and a domain guided dropout technique is proposed to improve the feature learning. Several recent works addressed on solving person re-id on abnormal images, such as low-resolution images~\cite{li2015multi}, or partially occluded images~\cite{zheng2015partial}.

Concurrent with our prior arXiv submission, Zheng~\etal~\cite{zheng2016person} also contributed a benchmark dataset for person search. They exploited separate detection and re-id methods with scores re-weighting to solve the problem, while in this work we propose a deep learning framework that jointly handles both aspects.

\textbf{Pedestrian detection.} DPM~\cite{felzenszwalb2010object}, ACF~\cite{dollar2014fast}, and Checkerboards~\cite{zhang2015filtered} are the most commonly used off-the-shelf pedestrian detectors. They rely on hand-crafted features and linear classifiers to detect pedestrians. Recent years, CNN-based pedestrian detectors have also been developed~\cite{yang2015convolutional,zhang2016faster}. Various factors, including CNN model structures, training data, and different training strategies are studied empirically in~\cite{hosang2015taking}. Tian~\etal~\cite{tian2015pedestrian} exploited pedestrian and scene attribute labels to train CNN pedestrian detectors in a multi-task manner. Cai~\etal~\cite{cai2015learning} proposed a complexity-aware boosting algorithm for learning CNN detector cascades.

\begin{figure*}[t]
\begin{center}
\includegraphics[width=0.9\linewidth]{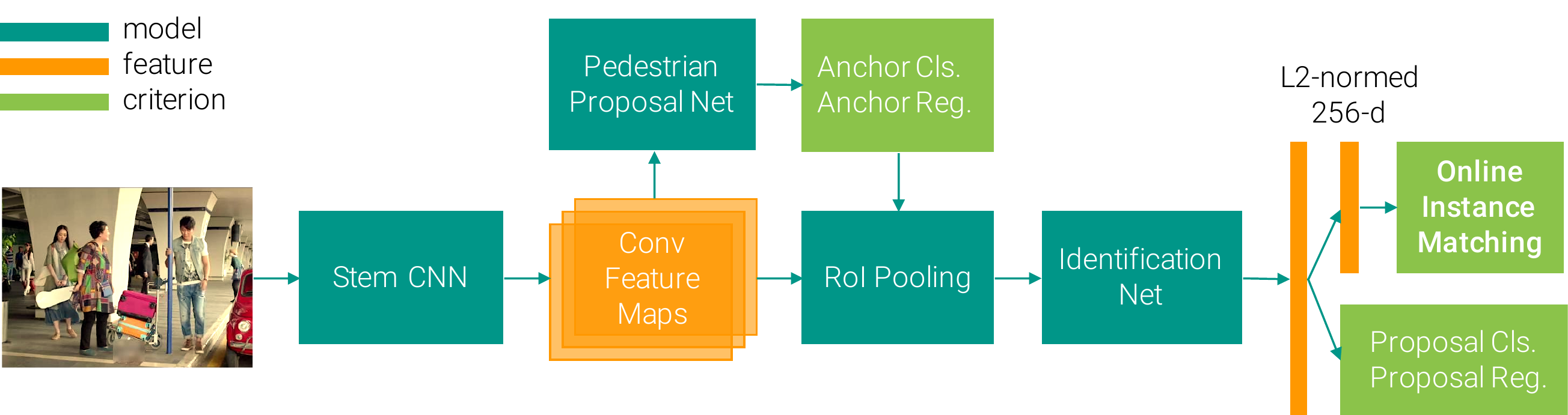}
\end{center}
\caption{Our proposed framework. Pedestrian proposal net generates bounding boxes of candidate people, which are fed into an identification net for feature extraction. We project the features to a L2-normalized $256$-d subspace, and train it with a proposed Online Instance Matching loss. Both the pedestrian proposal net and the identification net share the underlying convolutional feature maps.}
\label{fig:framework}
\end{figure*}
\section{Method} 
\label{sec:method}

We propose a new deep learning framework that jointly handles the pedestrian detection and person re-identification in a single convolutional neural network (CNN), as shown in Figure~\ref{fig:framework}. Given as input a whole scene image, we first use a stem CNN to transform from raw pixels to convolutional feature maps. A pedestrian proposal net is built upon these feature maps to predict bounding boxes of candidate people, which are then fed into an identification net with RoI-Pooling~\cite{girshick2015fast} to extract L2-normalized $256$-d features for each of them. At inference stage, we rank the gallery people according to their feature distances to the target person. At training stage, we propose an \textit{Online Instance Matching} (OIM) loss function on top of the feature vectors to supervise the identification net, together with several other loss functions for training the proposal net in a multi-task manner. Below we will first detail the CNN model structure, and then elaborate on the OIM loss function.

\subsection{Model Structure} 
\label{sub:model_structure}
We adopt the ResNet-50~\cite{he2015deep} as our base CNN model. It has a $7\times 7$ convolution layer in front (named conv1), followed by four blocks (named conv2\_x to conv5\_x) each containing $3, 4, 6, 3$ residual units, respectively. We exploit conv1 to conv4\_3 as the stem part. Given an input image, the stem will produce $1024$ channels of features maps, which have $1/16$ resolutions of the original image.

On top of these feature maps, we build a pedestrian proposal network to detect person candidates. A $512\times3\times3$ convolutional layer is first added to transform the features specifically for pedestrians. Then we follow~\cite{ren2015faster} to associate $9$ anchors at each feature map location, and use a Softmax classifier to predict whether each anchor is a pedestrian or not, as well as a linear regression to adjust their locations. We will keep the top $128$ adjusted bounding boxes after non-maximum suppression as our final proposals.

To find the target person among all these proposals, we build an identification net to extract the features of each proposal, and compare against the target ones. We first exploit an RoI-Pooling layer~\cite{girshick2015fast} to pool a $1024\times 14\times 14$ region from the stem feature maps for each proposal. Then they are passed through the rest conv4\_4 to conv5\_3 of the ResNet-50, followed by a global average pooling layer to summarize into a $2048$ dimensional feature vector. On one hand, as the pedestrian proposals would inevitably contain some false alarms and misalignments, we use again a Softmax classifier and a linear regression to reject non-persons and refine the locations. On the other hand, we project the features into a L2-normalized $256$ dimensional subspace (id-feat), and use them to compute cosine similarities with the target person when doing inference. During the training stage, we supervise the id-feat with the proposed OIM loss function. Together with other loss functions for detection, the whole net is jointly trained in a multi-task learning manner, rather than using the alternative optimizations in~\cite{ren2015faster}.

\subsection{Online Instance Matching Loss} 
\label{sub:online_instance_matching_loss}

\begin{figure*}[t]
\begin{center}
\includegraphics[width=0.9\linewidth]{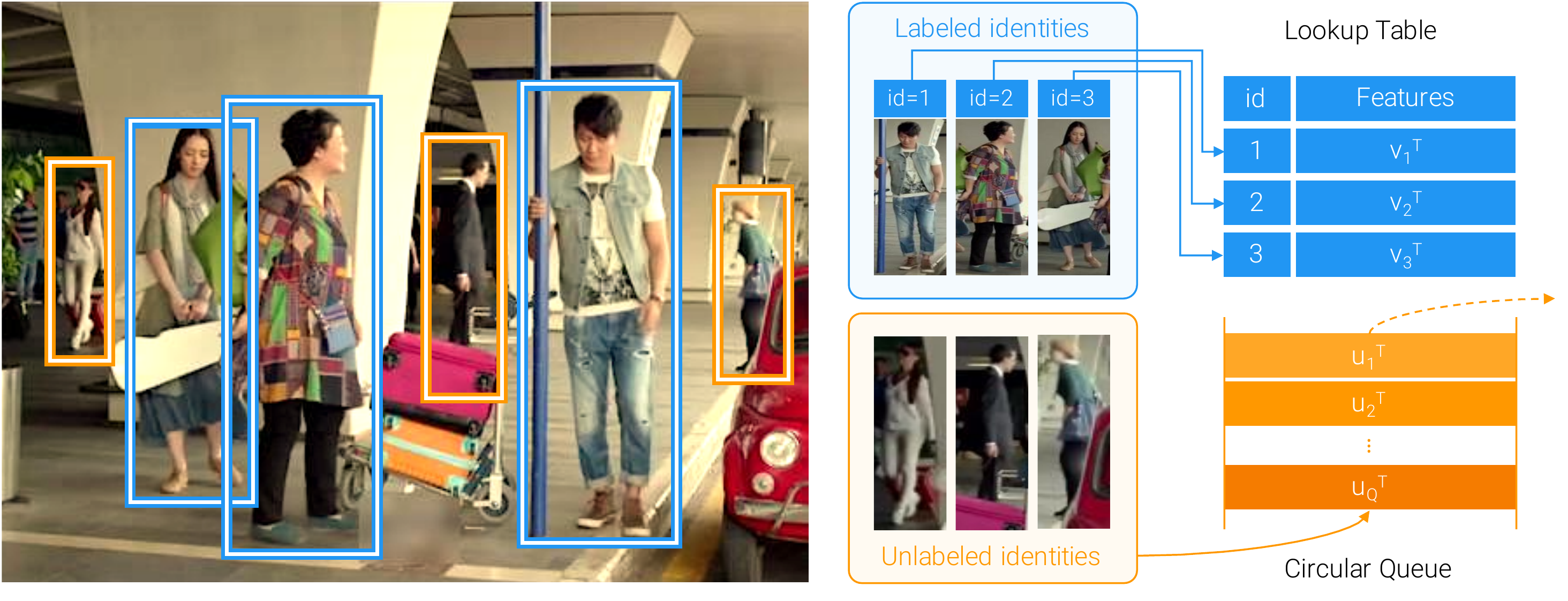}
\end{center}
\caption{Online Instance Matching. The left part shows the labeled (blue) and unlabeled (orange) identity proposals in an image. We maintain a lookup table (LUT) and a circular queue (CQ) to store the features. When forward, each labeled identity is matched with all the stored features. When backward, we update LUT according to the id, pushing new features to CQ, and pop out-of-date ones. Note that both data structures are external buffer, rather than the parameters of the CNN.}
\label{fig:oim}
\end{figure*}

There are three different types of proposals, labeled identities, unlabeled identities, and background clutter. Suppose there are $L$ different target people in the training set, when a proposal matches a target person, we call it an instance of the labeled identity, and assign a class-id (from $1$ to $L$) to it accordingly. There are also lots of proposals predicting pedestrians correctly, but do not belong to anyone of our target people. We call them unlabeled identities in such cases. We demonstrate some examples of labeled and unlabeled identities in Figure~\ref{fig:oim} with blue and orange bounding boxes, respectively. Other proposals are just false alarms on other objects or background regions. In the proposed loss function, we only consider the labeled and unlabeled identities, while leave the other proposals untouched.

As our goal is to distinguish different people, a natural objective is to minimize the features discrepancy among the instances of the same person, while maximize the discrepancy among different people. To fulfill this goal, we need to memorize the features of all the people. This could be done offline by doing network forward on all the training images, but it is not practical when using stochastic gradient descent (SGD) for optimization. Thus in our approach, we choose an online approximation instead. Denote the features of a labeled identity inside a mini-batch by $x\in \mathbb{R}^D$, where $D$ is the feature dimension, we maintain a lookup table (LUT) $V\in \mathbb{R}^{D\times L}$ to store the features of all the labeled identities, as demonstrated in Figure~\ref{fig:oim}. During the forward propagation, we compute cosine similarities between the mini-batch sample and all the labeled identities by $V^T x$. During backward, if the target class-id is $t$, then we will update the $t$-th column of the LUT by $v_t \gets \gamma v_t + (1-\gamma) x$, where $\gamma \in [0, 1]$, and then scale $v_t$ to have unit L2-norm.

Apart from labeled identities, many unlabeled identities are also valuable for learning feature representations. They can be safely used as negative classes for all the labeled identities. We use a circular queue to store the features of these unlabeled identities that appear in recent mini-batches. Denote the features in this circular queue by $U\in \mathbb{R}^{D\times Q}$, where $Q$ is the queue size, we can also compute their cosine similarities with the mini-batch sample by $U^Tx$. After each iteration, we push the new feature vectors into the queue, while pop the out-of-date ones to keep the queue size unchanged.

Based on these two data structures, we define the probability of $x$ being recognized as the identity with class-id $i$ by a Softmax function
\begin{equation} \label{eq:pi}
p_i=\frac{\exp(v_i^Tx/\tau)}{\sum_{j=1}^L\exp(v_j^Tx/\tau)+\sum_{k=1}^Q\exp(u_k^Tx/\tau)},
\end{equation}
where higher temperature $\tau$ leads to softer probability distribution. Similarly, the probability of being recognized as the $i$-th unlabeled identity in the circular queue is
\begin{equation} \label{eq:qi}
q_i=\frac{\exp(u_i^Tx/\tau)}{\sum_{j=1}^L\exp(v_j^Tx/\tau)+\sum_{k=1}^Q\exp(u_k^Tx/\tau)}.
\end{equation}
OIM objective is to maximize the expected log-likelihood
\begin{equation} \label{eq:loglik}
\mathcal{L}=\mathrm{E}_x\left[\log p_t\right],
\end{equation}
and its gradient with respect to $x$ can be derived as
\begin{equation}
\frac{\partial \mathcal{L}}{\partial x}=\frac{1}{\tau}\left[(1-p_t)v_t - \sum_{\substack{j=1\\j\ne t}}^L p_j v_j - \sum_{k=1}^Q q_k u_k\right].
\end{equation}

It can be seen that our OIM loss effectively compares the mini-batch sample with all the labeled and unlabeled identities, driving the underlying feature vector to be similar with the target one, while pushing it away from the others.

\textbf{Why not Softmax loss?} A natural question here is that why not learning a classifier matrix with a conventional Softmax loss to predict the class-id. There are mainly two drawbacks. First, large-scale person search datasets would have a large number of identities (more than $5,000$ in our training set), while each identity only has several instances and each image only contains a few identities. We need to learn more than $5,000$ discriminant functions simultaneously, but during each SGD iteration we only have positive samples from tens of classes. The classifier matrix suffers from large variance of gradients and thus cannot be learned effectively, even with proper pre-training and high momentum. Second, we cannot exploit the unlabeled identities with Softmax loss, as they have no specific class-ids.

Although our OIM loss formulation is similar to the Softmax one, the major difference is that the OIM loss is non-parametric. The LUT and circular queue are considered as external buffer, rather than the network parameters. The gradients directly operate on the features without the transformation by a classifier matrix. The potential drawback of this non-parametric loss is that it could overfit more easily. We find that projecting the features into a L2-normalized low-dimensional subspace helps reduce overfitting.

\textbf{Scalability.} Computing the partition function in Eq~\eqref{eq:pi} and Eq~\eqref{eq:qi} could be time consuming when the number of identities increases. To overcome this problem, we can approximate the denominators by sub-sampling the labeled and unlabeled identities, which results in optimizing a lower-bound of Eq~\eqref{eq:loglik}.


\section{Dataset} 
\label{sec:dataset}
We collect and annotate a large-scale person search dataset to evaluate of our proposed method. We exploit two data sources to diversify the scenes. On one hand, we use hand-held cameras to shoot street snaps around an urban city. On the other hand, we collect from movie snapshots that contain pedestrians, as they could enrich the variations of viewpoints, lighting, and background conditions. In this section, we will show the basic statistics of our dataset, as well as define the evaluation protocols and metrics.

\subsection{Statistics} 
\label{sub:statistics}
\begin{figure}[t]
\begin{center}
\includegraphics[width=0.75\linewidth]{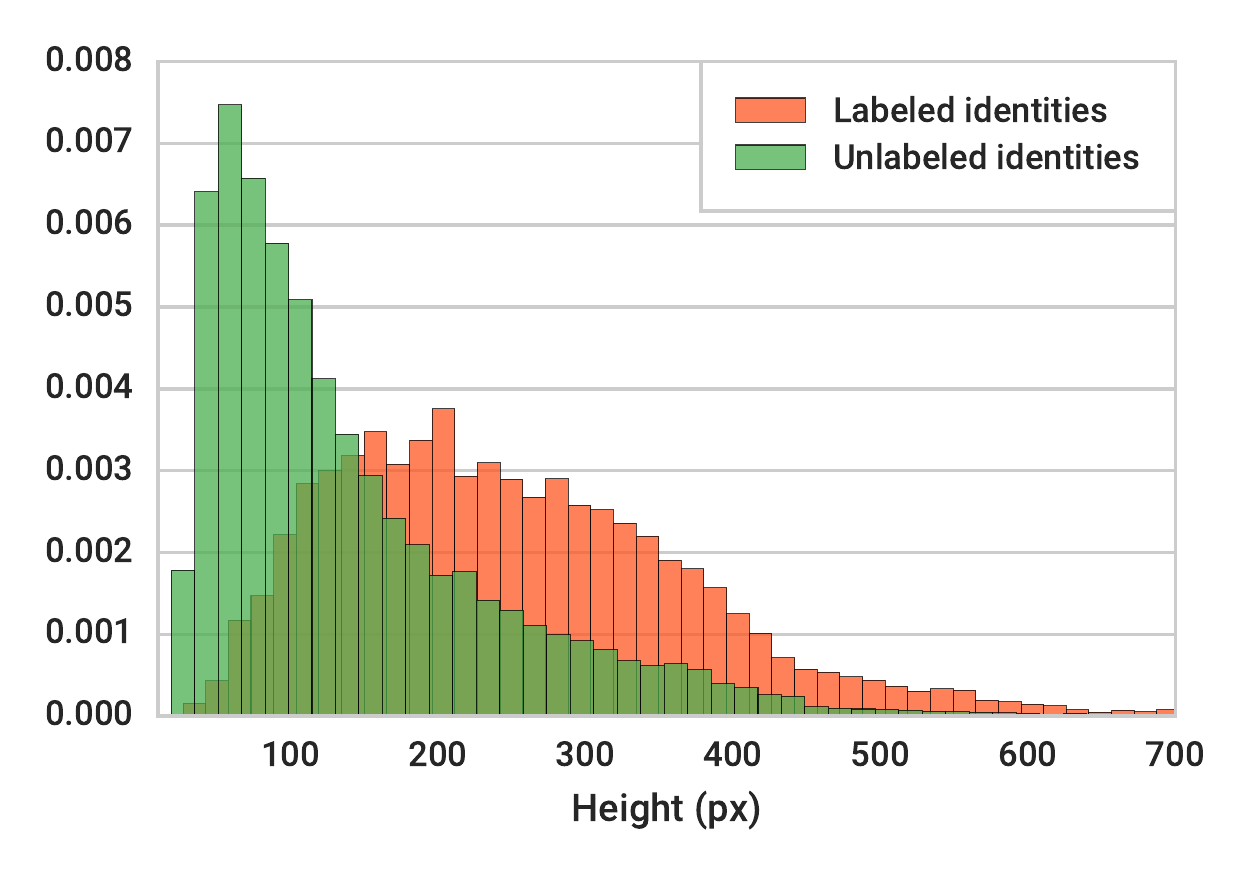}
\end{center}
\vspace{-3ex}
\caption{The height distributions of labeled and unlabeled identities in our dataset.}
\vspace{-1ex}
\label{fig:height-distr}
\end{figure}

After collecting all the $18,184$ images, we first densely annotate all the $96,143$ pedestrians bounding boxes in these scenes, and then associate the person that appears across different images, resulting in $8,432$ labeled identities. The statistics of two data sources are listed in Table~\ref{tab:dataset-stats}. We did not annotate those people who appear with half bodies or abnormal poses such as sitting or squatting. Moreover, people who change clothes and decorations in different video frames are not associated in our dataset, since person search problem requires to recognize identities mainly according to their clothes and body shapes rather than faces. We ensure that the background pedestrians do not contain labeled identities, and thus they can be safely served as negative samples for identification. Note that we also ignore the background pedestrians whose heights are smaller than $50$ pixels, as they would be hard to recognize even for human labelers. The height distributions of labeled and unlabeled identities are demonstrated in Figure~\ref{fig:height-distr}. It can be seen that our dataset has rich variations of pedestrian scales.
\begin{table}
\small
\begin{center}
\begin{tabular}{lrrr}
\hline\noalign{\smallskip}
\noalign{\smallskip}
Source / Split & \# Images & \# Pedestrians & \# Identities \\
\noalign{\smallskip}\hline\hline\noalign{\smallskip}
StreetSnap & 12,490 & 75,845 & 6,057 \\
Movie\&TV & 5,694 & 20,298 & 2,375 \\
\hline\noalign{\smallskip}
Training & 11,206 & 55,272 & 5,532 \\
Test & 6,978 & 40,871 & 2,900 \\
\hline\noalign{\smallskip}
Overall & 18,184 & 96,143 & 8,432 \\
\hline\noalign{\smallskip}
\end{tabular}
\end{center}
\vspace{-3ex}
\caption{Statistics of the dataset with respect to data sources and training / test splits.}
\label{tab:dataset-stats}
\end{table}

\subsection{Evaluation Protocols and Metrics} 
\label{sub:evaluation_protocols_and_metrics}
We split the dataset into a training and a test subset, ensuring no overlapped images or labeled identities between them. Table~\ref{tab:dataset-stats} shows the statistics of these two subsets. We divide the test identity instances into queries and galleries. For each of the $2,900$ test identities, we randomly choose one of his/her instances as the query, while the corresponding gallery set consists of two parts---all the images containing the other instances and some randomly sampled images not containing this person. Different queries have different galleries, and jointly they cover all the $6,978$ test images.

To better understand how gallery size would affect the person search performance, we define a set of protocols with gallery size ranging from $50$ to $4000$. Taking gallery size of $100$ as an example, as each image approximately contains $6$ pedestrians, then our task is to find the target person among about $600$ people. This setting is comparable with existing person re-id datasets (\eg, CUHK-03, VIPeR) in terms of the number of gallery pedestrians, and is even more challenging as there could be thousands of background clutter bounding boxes distracting our attentions.

We employ two kinds of evaluation metrics---cumulative matching characteristics (CMC top-K) and mean averaged precision (mAP). The first one is inherited from the person re-id problem, where a matching is counted if there is at least one of the top-K predicted bounding boxes overlaps with the ground truths with intersection-over-union (IoU) greater or equal to $0.5$. The second one is inspired from the object detection tasks. We follow the ILSVRC object detection criterion~\cite{russakovsky2014imagenet} to judge the correctness of predicted bounding boxes. An averaged precision (AP) is calculated for each query based on the precision-recall curve, and then we average the APs across all the queries to get the final result.

\section{Experiments} 
\label{sec:experiments}
To evaluate the effectiveness of our approach and study the impact of various factors on person search performance, we conduct several groups of experiments on the new dataset. In this section, we first detail the baseline methods and experiment settings in Section~\ref{sub:experiment_settings}. Then we compare our joint framework with the baselines of using separate pedestrian detection and person re-identification in Section~\ref{sub:comparison_with_detection_and_re_id}. Section~\ref{sub:effectiveness_of_online_instance_matching} shows the effectiveness of our proposed Online Instance Matching (OIM) loss. At last, we present the influence of various factors, including detection recall and gallery size.

\subsection{Experiment Settings} 
\label{sub:experiment_settings}
We implement our framework based on Caffe~\cite{jia2014caffe,wang2016temporal} and py-faster-rcnn~\cite{girshick2015fast,ren2015faster}. ImageNet-pretrained ResNet-50~\cite{he2015deep} are exploited for parameters initialization. We fix the first $7\times 7$ convolution layer and the batch normalization (BN) layers as constant affine transformations in the stem part, while keep the other BN layers as normal in the identification part. The temperature scalar $\tau$ in Eq.~\eqref{eq:pi} and Eq.~\eqref{eq:qi} is set to $0.1$, the size of the circular queue is set to $5,000$. All the losses have the same loss weight. Each mini-batch consists of two scene images. The learning rate is initialized to $0.001$, dropped to $0.0001$ after $40$K iterations, and kept unchanged until the model converges at $50$K iterations.

\begin{figure}[t]
\begin{center}
\includegraphics[width=0.75\linewidth]{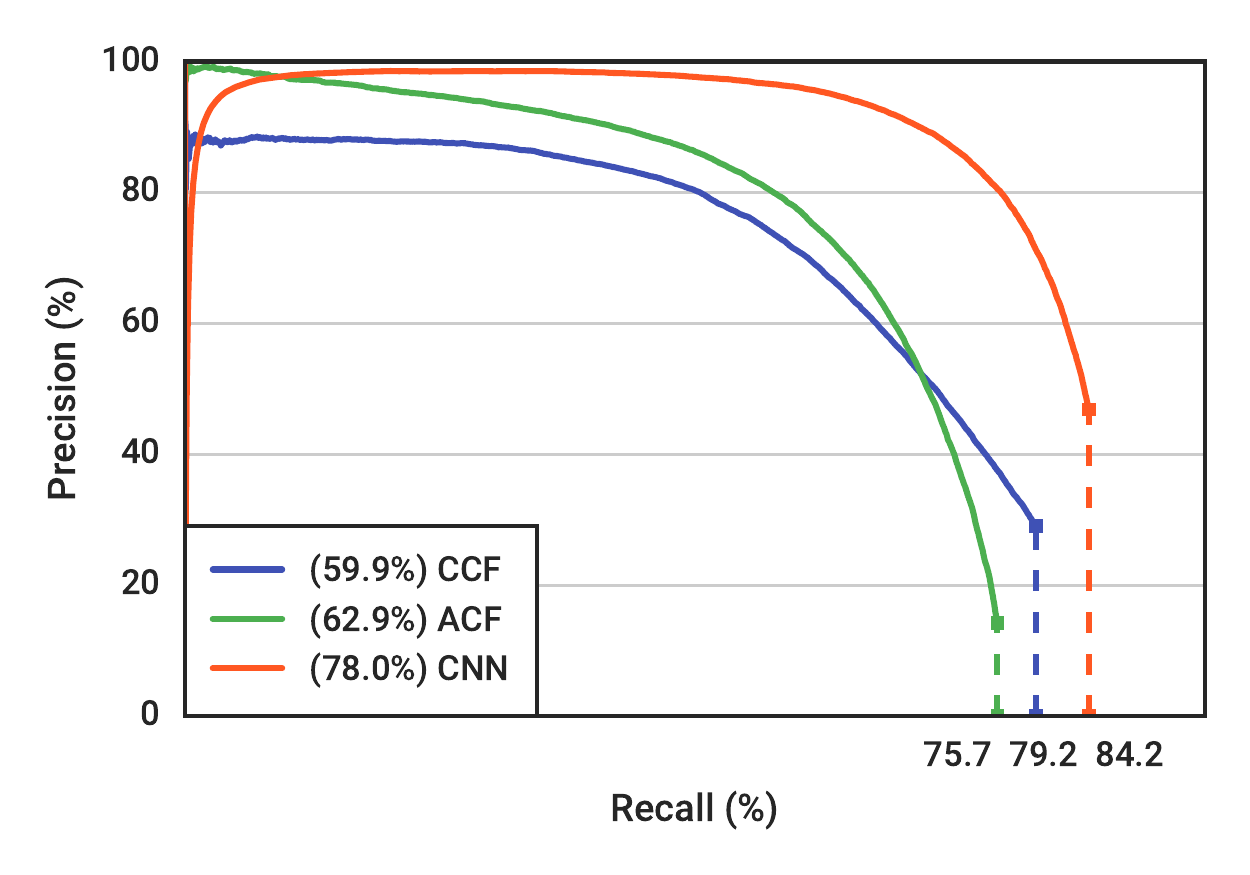}
\end{center}
\vspace{-4ex}
\caption{Recall-Precision curves of different detectors. APs are listed in the legend.}
\label{fig:detector-pr-curves}
\end{figure}

We compare our framework with conventional methods that break down the problem into two separate tasks---pedestrian detection and person re-identification. Three pedestrian detection and five person re-id methods are used in our experiments, resulting in $15$ baseline combinations. For pedestrian detection, we directly use the off-the-shelf deep learning CCF~\cite{yang2015convolutional} detector, as well as two other detectors specifically fine-tuned on our dataset. One is the ACF~\cite{dollar2014fast}, and the other is Faster-RCNN (CNN)~\cite{ren2015faster} with ResNet-50, which is equivalent to our framework but without the identification task. The recall-precision curve of each detector on our dataset are plotted in Figure~\ref{fig:detector-pr-curves}. We also use the ground truth (GT) bounding boxes as the results of a perfect detector.

For person re-identification, we use several popular re-id feature representations, including DenseSIFT-ColorHist (DSIFT)~\cite{zhao2013unsupervised}, Bag of Words (BoW)~\cite{zheng2015scalable}, and Local Maximal Occurrence (LOMO)~\cite{liao2015person}. Each feature representation is used in conjunction with a specific distance metric, including Euclidean, Cosine similarity, KISSME~\cite{koestinger2012large}, and XQDA~\cite{liao2015person}, where KISSME and XQDA are trained on our dataset. Moreover, by discarding the pedestrian proposal network in our framework and training the remaining net to classify identities with Softmax loss from cropped pedestrian images, we get another baseline re-id method (IDNet). This training scheme has been exploited in~\cite{xiao2016learning} to learn discriminative re-id feature representations. In our experiments, when training IDNet with detector boxes, we found that adding background clutter as a unique class improves the result, while adding unlabeled identities does not.

The following results are reported using the protocol with gallery size equal to $100$ if not specified.

\subsection{Comparison with Detection and Re-ID} 
\label{sub:comparison_with_detection_and_re_id}
We first compare our proposed person search framework (with or without using unlabeled identities) with other $15$ baseline combinations that break down the problem into separate detection and re-identification tasks. The results are summarized in Table~\ref{tab:results}. Our method outperforms the others by large margin. Comparing with CNN+IDNet, the gain comes from the joint optimization of the detection and identification parts, as well as the effective use of unlabeled identities in the OIM loss.

\setlength{\tabcolsep}{8pt}
\begin{table}
\small
\begin{center}
\begin{tabular}{lrrrr}
\hline\noalign{\smallskip}
\noalign{\smallskip}
\multicolumn{1}{c}{CMC top-1 (\%)} & CCF & ACF & CNN & GT \\
\noalign{\smallskip}\hline\hline\noalign{\smallskip}
DSIFT+Euclidean & 11.7 & 25.9 & 39.4 & 45.9 \\
DSIFT+KISSME    & 13.9 & 38.1 & 53.6 & 61.9 \\
BoW+Cosine      & 29.3 & 48.4 & 62.3 & 67.2 \\
LOMO+XQDA       & 46.4 & 63.1 & 74.1 & 76.7 \\
IDNet           & 57.1 & 63.0 & 74.8 & 78.3 \\
\hline\noalign{\smallskip}
Ours (w/o unlabeled)   & --- & --- & 76.1 & 78.5 \\
Ours                   & --- & --- & \textbf{78.7} & \textbf{80.5} \\
\hline\noalign{\smallskip}
\noalign{\smallskip}
\multicolumn{1}{c}{mAP (\%)} & CCF & ACF & CNN & GT \\
\noalign{\smallskip}\hline\hline\noalign{\smallskip}
DSIFT+Euclidean & 11.3 & 21.7 & 34.5 & 41.1 \\
DSIFT+KISSME    & 13.4 & 32.3 & 47.8 & 56.2 \\
BoW+Cosine      & 26.9 & 42.4 & 56.9 & 62.5 \\
LOMO+XQDA       & 41.2 & 55.5 & 68.9 & 72.4 \\
IDNet           & 50.9 & 56.5 & 68.6 & 73.1 \\
\hline\noalign{\smallskip}
Ours (w/o unlabeled)   & --- & --- & 72.7 & 75.5 \\
Ours                   & --- & --- & \textbf{75.5} & \textbf{77.9} \\
\hline\noalign{\smallskip}
\end{tabular}
\end{center}
\vspace{-3ex}
\caption{Comparisons between our framework and separate pedestrian detection + person re-id methods. }
\vspace{-1ex}
\label{tab:results}
\end{table}
\setlength{\tabcolsep}{6pt}

From Table~\ref{tab:results} we can also see that different detectors affect the person search performance significantly for each re-id method. Directly using an off-the-shelf detector may not be a good choice when applying existing re-id methods in the real-world person search applications. Otherwise the detector could become a bottleneck that diminishes the returns of better re-id methods.

On the other hand, the relative performance of different re-id methods are consistent across all the detectors. It implies that existing person re-id datasets could still guide us to design better feature representations, but it may lose some valuable data, such as unlabeled identities and background clutter, which come with the person search datasets.

Another interesting phenomenon is that although IDNet and LOMO+XQDA have similar performance when using GT or fine-tuned ACF and CNN detectors, IDNet is significantly better when using off-the-shelf CCF detector. We observe that the CCF detection results contain many misalignments. Hand-crafted features in such cases are not as robust as the IDNet counterpart.

\subsection{Effectiveness of Online Instance Matching} 
\label{sub:effectiveness_of_online_instance_matching}
\begin{figure}[t]
\begin{center}
\includegraphics[width=0.8\linewidth]{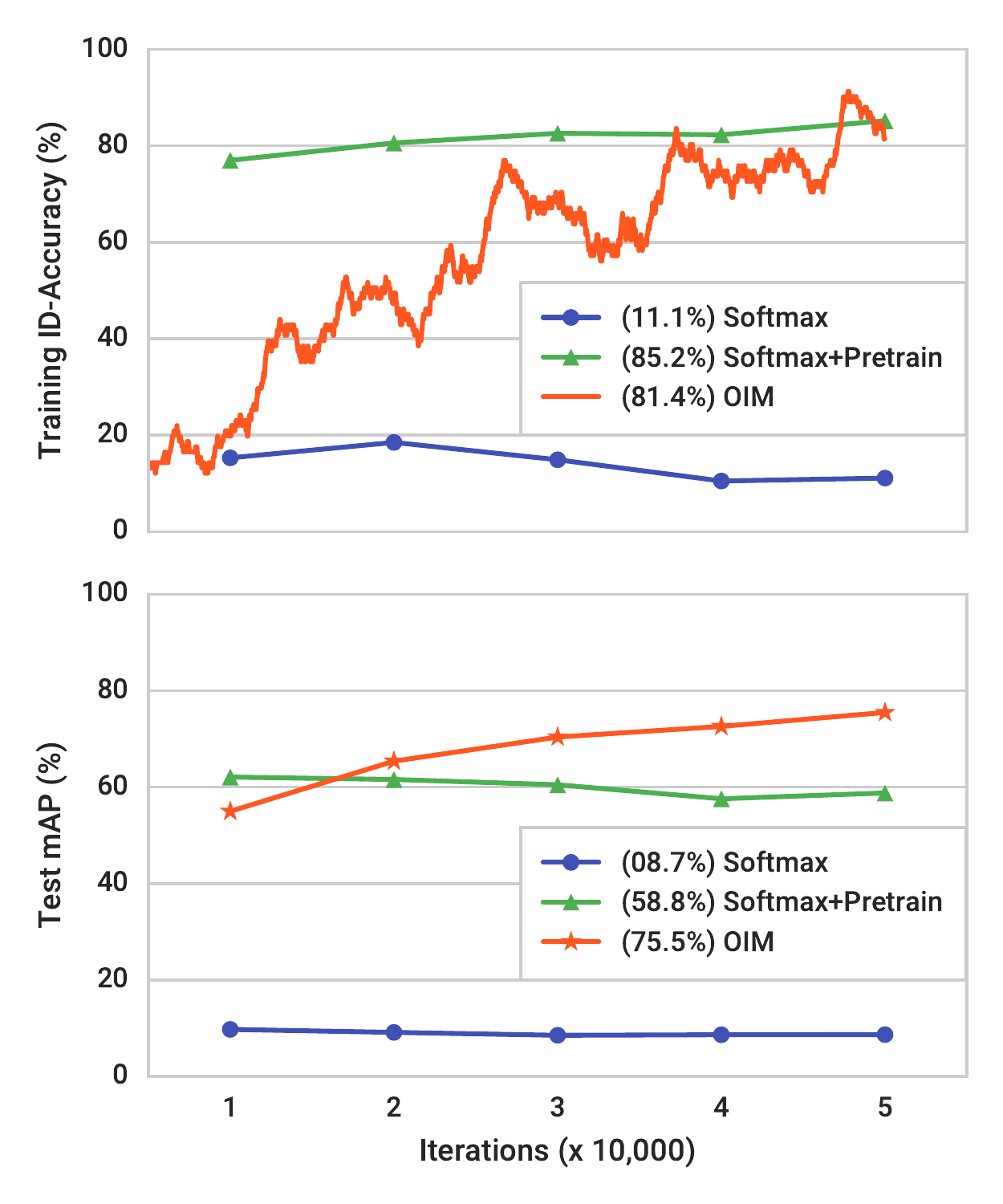}
\end{center}
\vspace{-3ex}
\caption{Comparisons between using the proposed Online Instance Matching (OIM) and Softmax loss (with and without pretraining the Softmax classifier) in our framework. The final accuracies and mAPs are shown in the legends.}
\vspace{1ex}
\label{fig:oim-vs-softmax}
\end{figure}
We validate the effectiveness of the proposed Online Instance Matching (OIM) loss by comparing it against Softmax baselines with or without pretraining the classifier matrix. The training identification accuracy and test person search mAP curves are demonstrated in Figure~\ref{fig:oim-vs-softmax}. First, we can see that using Softmax loss without pretraining classifier remains at low accuracy during the whole process. This phenomenon verifies our analysis in Section~\ref{sub:online_instance_matching_loss} that learning a large classifier matrix is difficult. Even with proper pretraining, the training accuracy still improves slowly, and the test mAP keeps at around $60\%$.

On the contrary, the proposed OIM loss starts with a low training accuracy but converges much faster and also consistently improves the test performance. The parameter-free OIM loss learns features directly without needing to learn a big classifier matrix. Moreover, the mismatch between training and test criterion no longer exists, as both are computed based on the inner product of L2-normalized feature vectors, which represents the cosine similarity.

We further evaluate the impact of OIM loss on the standard person re-identification task. We train two different base CNNs, Inception~\cite{xiao2016learning} (from scratch) and ResNet-50~\cite{he2015deep} (ImageNet pretrained), with either Softmax loss or OIM loss, on three large-scale person re-id datasets, CUHK03~\cite{li2014deepreid}, Market1501~\cite{zheng2015scalable}, and Duke~\cite{zheng2017unlabeled,ristani2016MTMC}. Following their own protocols, we evaluate the CMC top-1 accuracy of using different loss functions, as listed in Table~\ref{tab:oim-reid}. OIM loss consistently outperforms Softmax loss, regardless of which base CNN is used. We refer readers to Open-ReID\footnote{\url{https://github.com/Cysu/open-reid}} benchmarks for more details.
\setlength{\tabcolsep}{8pt}
\begin{table}
\small
\begin{center}
\begin{tabular}{llccc}
\hline\noalign{\smallskip}
\noalign{\smallskip}
Network & Loss & CUHK03 & Market1501 & Duke \\
\noalign{\smallskip}\hline\hline\noalign{\smallskip}
Inception & Softmax & 73.2 & 75.8 & 54.4 \\
Inception & OIM & \textbf{77.7} & \textbf{77.9} & \textbf{61.7} \\
\hline\noalign{\smallskip}
ResNet-50 & Softmax & 70.8 & 81.4 & 62.5 \\
ResNet-50 & OIM & \textbf{77.5} & \textbf{82.1} & \textbf{68.1} \\
\hline\noalign{\smallskip}
\end{tabular}
\end{center}
\vspace{-3ex}
\caption{CMC top-1 accuracy (\%) of using Softmax or OIM loss for standard person re-id task.}
\vspace{-1ex}
\label{tab:oim-reid}
\end{table}
\setlength{\tabcolsep}{6pt}

\begin{figure*}[t]
\begin{center}
\begin{subfigure}[t]{0.32\linewidth}
  \includegraphics[width=\linewidth]{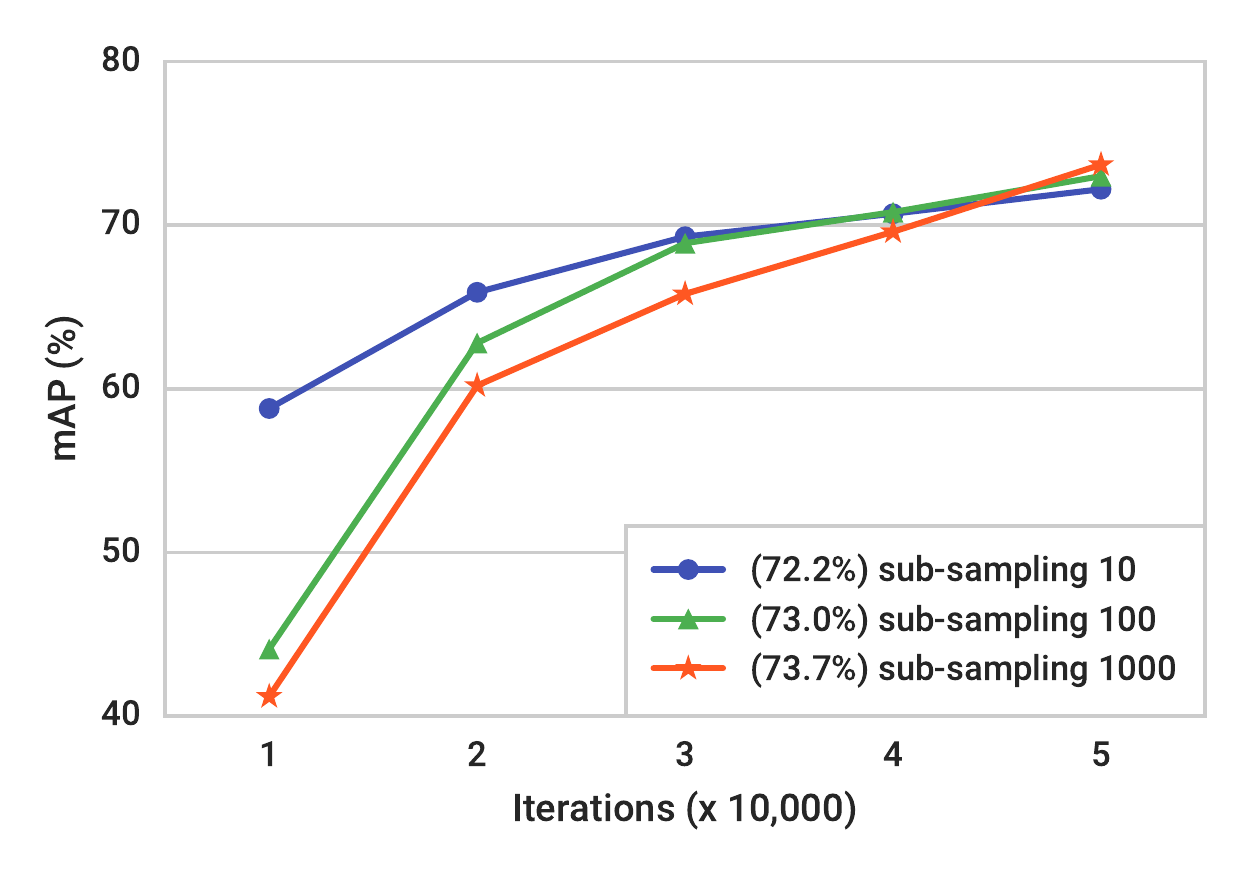}
  \caption{Sub-sampling size for the OIM loss}
  \label{fig:subsampling}
\end{subfigure}
~
\begin{subfigure}[t]{0.32\linewidth}
  \includegraphics[width=\linewidth]{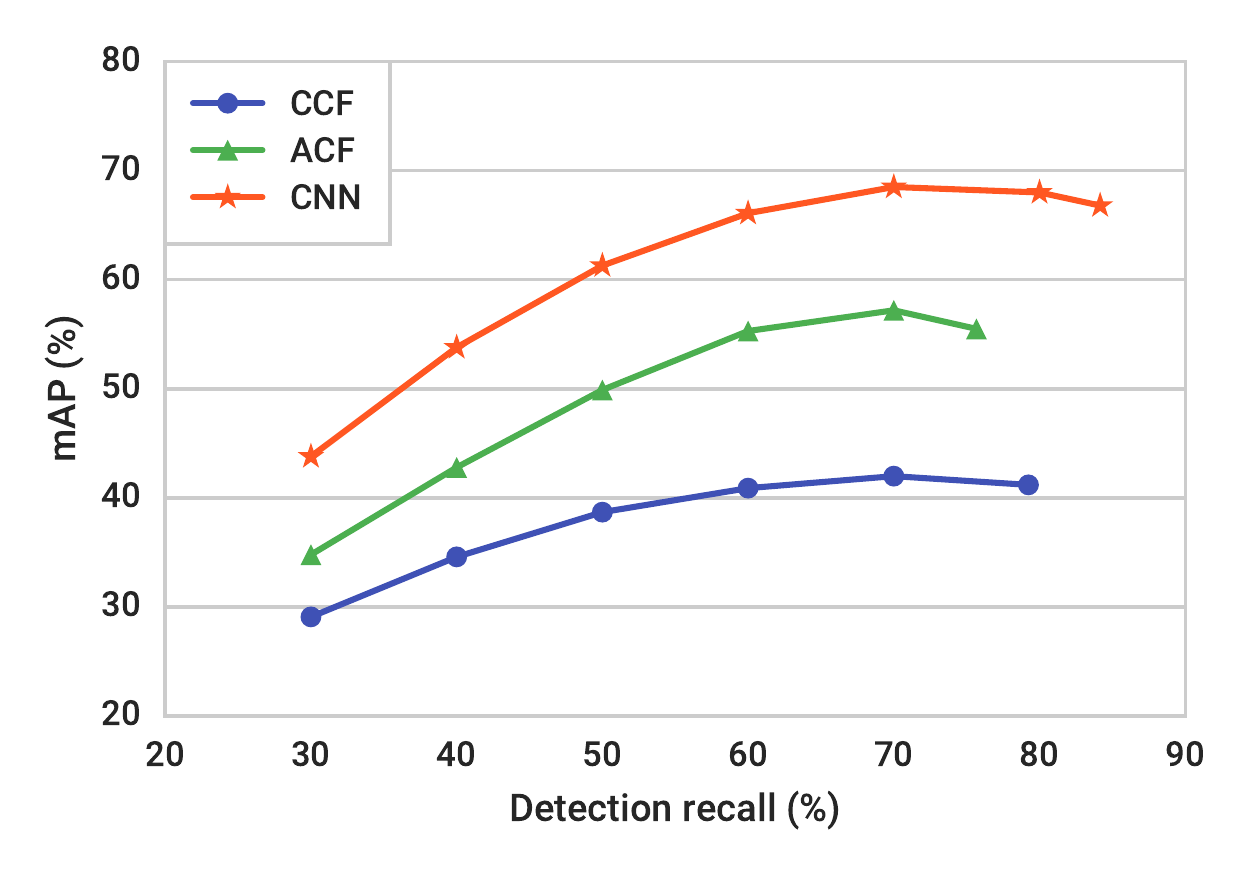}
  \caption{Detection (LOMO+XQDA) recall}
  \label{fig:factor-recall}
\end{subfigure}
~
\begin{subfigure}[t]{0.32\linewidth}
  \includegraphics[width=\linewidth]{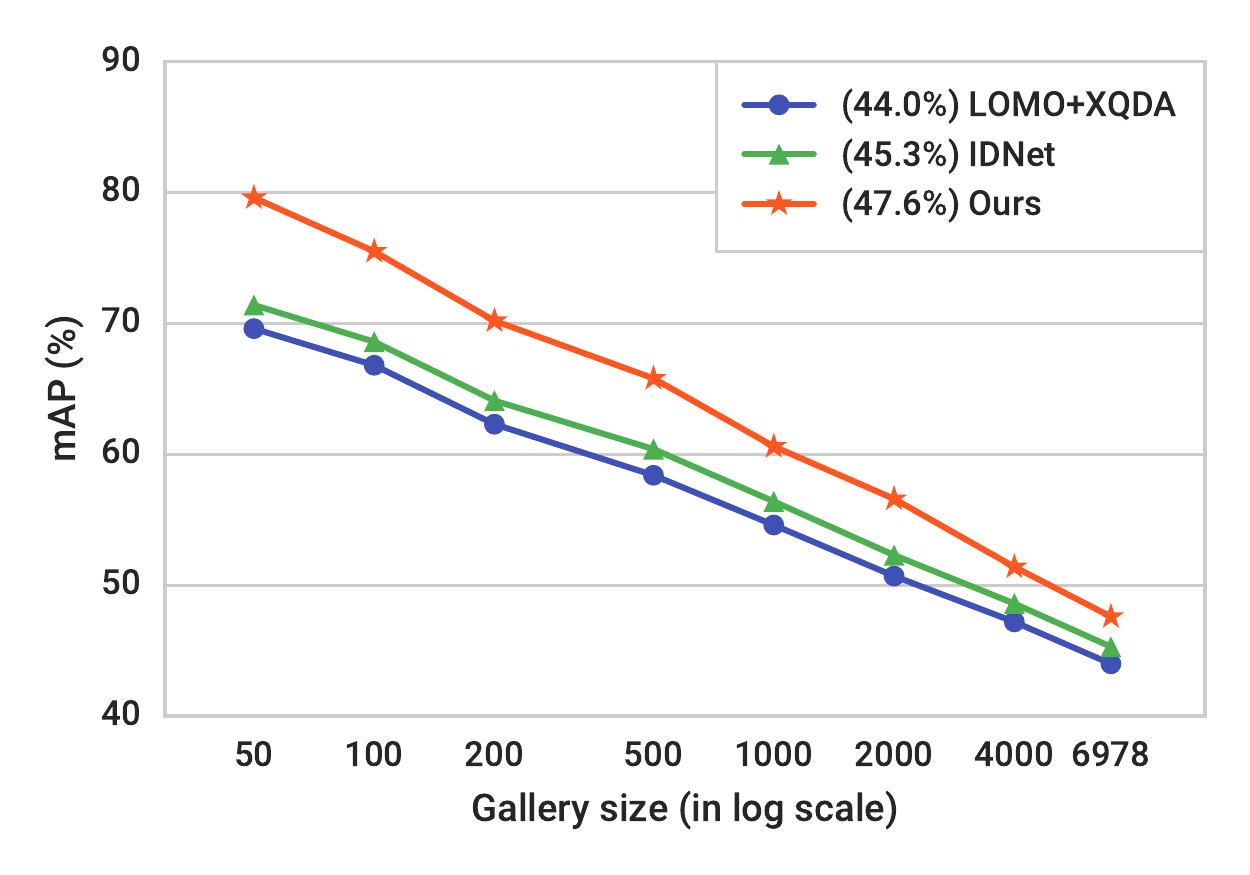}
  \caption{Test gallery size}
  \label{fig:factor-gallery-size}
\end{subfigure}
\end{center}
\caption{Test mAP curves of different factors. The final mAPs are shown in the legend if applicable.}
\end{figure*}
\textbf{Sub-sampling the identities.} As the number of identities increases, the computation time of the OIM loss could become the bottleneck of the whole system. Thus we proposed in Section~\ref{sub:online_instance_matching_loss} to approximate Eq.~\eqref{eq:pi} and Eq.~\eqref{eq:qi} by sub-sampling both the labeled and unlabeled identities in the denominators. We validate this approach here by training the framework with sub-sampling size of $10$, $100$, and $1000$. The test mAP curves are demonstrated in Figure~\ref{fig:subsampling}. In general, sub-sampling a small number of identities relaxes the training objective, which leads to slightly inferior performance but much faster convergence rate. This indicates that our framework is scalable to larger datasets with even more identities by using proper sub-sampling rate.

\setlength{\tabcolsep}{8pt}
\begin{table}
\small
\begin{center}
\begin{tabular}{lrrrrr}
\hline\noalign{\smallskip}
\noalign{\smallskip}
Dimension & N/A & 128 & 256 & 512 & 1024 \\
\noalign{\smallskip}\hline\hline\noalign{\smallskip}
top-1 (\%) & 59.3 & 65.9 & \textbf{78.7} & 78.2 & 78.5 \\
mAP (\%) & 54.2 & 62.1 & 75.5 & 75.3 & \textbf{75.7} \\
\hline\noalign{\smallskip}
\end{tabular}
\end{center}
\vspace{-3ex}
\caption{Comparisons among different dimensions of L2-normalized feature subspace. N/A means that we directly use the L2-normalized $2048$-d global pooled feature vector.}
\vspace{-1ex}
\label{tab:embed-dim}
\end{table}
\setlength{\tabcolsep}{6pt}

\textbf{Low-dimensional subspace.} We further investigate how the dimension of the L2-normalized feature vector affects the person search performance. The results are summarized in Table~\ref{tab:embed-dim}. We observe that using the $2048$-d global pooled feature vector directly with L2-normalization leads to lower training error, but its test performance is $20\%$ worse. This suggests that projecting the features into a proper low-rank subspace is very important to regularize the network training. In our experiments, $256$ to $1024$ dimensions have similar test performance, and we choose $256$-d to accelerate the computation of feature distances.

\subsection{Factors for Person Search} 
\label{sub:factors_for_person_search}
\textbf{Detection Recall.} We investigate how detection recalls would affect the person search performance by using LOMO+XQDA as the re-id method and setting different thresholds on detection scores. A lower threshold reduces misdetections (increases the recall) but results in more false alarms. We choose the recall rates ranging from $30\%$ to the maximum value of each detector. The final person search mAP under each setting is demonstrated in Figure~\ref{fig:factor-recall}. An interesting observation is that higher recall does not necessarily lead to higher person search performance, which means re-id method could still get confused on some false alarms. This again indicates that we should not focus solely on training re-id methods with manually cropped pedestrians, but should consider the detections jointly under the person search problem setting.

\textbf{Gallery size.} Person search could be more challenging as the gallery size increases. We evaluate several methods under different test gallery sizes from $50$ to full set of $6,978$ images, following the protocols defined in Section~\ref{sub:evaluation_protocols_and_metrics}. The test mAPs are demonstrated in Figure~\ref{fig:factor-gallery-size}. Note that for each test query, the corresponding gallery images are randomly sampled from the whole set. All test images are covered even with small gallery sizes. The performance gaps among different methods are reduced as the gallery size increases, indicating all the methods may suffer from some common hard samples, and we could further improve the performance with hard example minings.

\section{Conclusion} 
\label{sec:conclusion}
In this paper, we propose a new deep learning framework for person search. It jointly handles detection and identification in a single CNN. An Online Instance Matching loss function is proposed to train the network effectively. Its non-parametric nature enables faster yet better convergence, which is validated through series of experiments.\\

\noindent\textbf{Acknowledgements.} This work is supported in part by SenseTime Group Limited, in part by the General Research Fund through the Research Grants Council of Hong Kong under Grants CUHK14213616, CUHK14206114, CUHK14205615, CUHK14207814, CUHK14203015, CUHK14239816, and CUHK419412, in part by the Hong Kong Innovation and Technology Support Programme Grant ITS/121/15FX, in part by the State Key Development Program under Grant 2016YFB1001004, in part by the Guangdong Natural Science Foundation under Grant No. 2014A030313201, and in part by National Natural Science Foundation of China under Grant 61371192.

{\small
\bibliographystyle{ieee}
\bibliography{1262}
}

\end{document}